\documentclass[10pt,journal,compsoc]{IEEEtran}

\ifCLASSOPTIONcompsoc

  \usepackage[nocompress]{cite}
\else
  \usepackage{cite}
\fi

\ifCLASSINFOpdf
  \usepackage[pdftex]{graphicx}
\else
\fi

\usepackage{amsmath}

\usepackage{csquotes}
\usepackage{color}

\usepackage{multirow}

\usepackage{url}

\begin{document}

\title{Content-Based Features to Rank Influential Hidden Services of the Tor Darknet}

\author{Mhd~Wesam~Al-Nabki,
    Eduardo~Fidalgo,
    Enrique~Alegre,
    and Deisy~Chaves
\IEEEcompsocitemizethanks{

\IEEEcompsocthanksitem M.~Al-Nabki, E.~Fidalgo, E.~Alegre, and D.~Chaves are with the School of Industrial Engineering, Computer Science and Aeronautics, Universidad de Le\'on, Spain. Also, they are Researchers at INCIBE (Spanish National Cybersecurity Institute), Le\'on, Spain. (e-mails:\{mnab, efidf, ealeg, d.chas\}@unileon.es).

}

}

\IEEEtitleabstractindextext{%
\begin{abstract}
The unevenness importance of criminal activities in the onion domains of the Tor Darknet and the different levels of their appeal to the end-user make them tangled to measure their influence. To this end, this paper presents a novel content-based ranking framework to detect the most influential onion domains. Our approach comprises a modeling unit that represents an onion domain using forty features extracted from five different resources: user-visible text, HTML markup, Named Entities, network topology, and visual content. And also, a ranking unit that, using the Learning-to-Rank (LtR) approach, automatically learns a ranking function by integrating the previously obtained features. Using a case-study based on drugs-related onion domains, we obtained the following results. (1) Among the explored LtR schemes, the listwise approach outperforms the benchmarked methods with an NDCG of 0.95 for the top-10 ranked domains. (2) We proved quantitatively that our framework surpasses the link-based ranking techniques. Also, (3) with the selected feature, we observed that the textual content, composed by text, NER, and HTML features, is the most balanced approach, in terms of efficiency and score obtained. The proposed framework might support Law Enforcement Agencies in detecting the most influential domains related to possible suspicious activities.

\end{abstract}

}

\maketitle

\IEEEdisplaynontitleabstractindextext

%
\IEEEpeerreviewmaketitle

\IEEEraisesectionheading{\section{Introduction}\label{sec:introduction}}

\IEEEPARstart{T}{he} Onion Router (Tor) network, which is known to be one of the most famous Darknet networks, gives the end-users a high level of privacy and anonymity. The Tor project was proposed in the mid-1990s by the US military researchers to secure intelligence communications. However, a few years later, and as part of their strategy for secrecy, they made the Tor project available for the public \cite{the_tor_project_2018}. 

The onion domains proliferated rapidly and the latest statistics stated by the onion metrics website\footnote{\url{https://metrics.torproject.org/hidserv-dir-onions-seen.html}} has reported that the number of currently existing onion domains has increased from $30$K to almost $90$K between April 2015 and October 2019.

The community of the Tor network refers to an onion domain hosted in the Tor darknet by a Hidden Service (HS). Those services can be accessed via a particular web browser called \textit{Tor Browser}\footnote{\url{https://www.torproject.org/projects/torbrowser.html.en}} or a proxy such as \textit{Tor2Web}\footnote{\url{https://tor2web.org/}}.

There are many legal uses for the Tor network, such as personal blogs, news domains, and discussion forums \cite{alnabki2019torank, choshen2019language}. However, due to its level of anonymity, Tor Darknet is being exploited by services traders allowing them to promote their products freely, including, but not limited to Child Sexual Abuse (CSA) \cite{Gangwar2017, foley2019sex}, drugs trading \cite{NABKI2017, foley2019sex, Ciancaglini2015, norbutas2018offline}, and counterfeit personal identifications \cite{Ling2015, RubelBiswasEduardoFidalgo2017, al2017classifying}. 

\begin{figure}[tp]
\centering
\includegraphics[width=\linewidth]{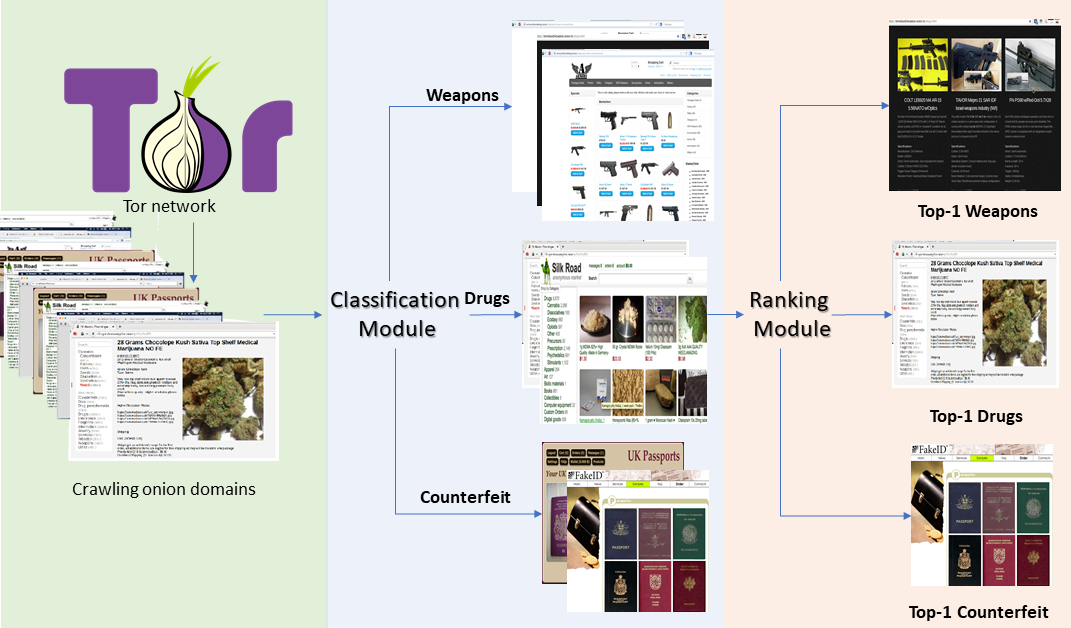}
\caption{Overview of a Tor monitoring system. Two of our previous research contributions are highlighted in orange.}
\label{fig:system_arch}
\end{figure}

The high level of privacy and anonymity provided by the Tor network obstructed the authorities monitoring tools from controlling the content or even identifying the IP address of the hosts who are behind any suspicious service.
We collaborate with the Spanish National Cybersecurity Institute (INCIBE\footnote{In Spanish, it stands for the Instituto Nacional de Ciberseguridad de Espa\~na}), to develop tools that could ease the task of monitoring the Tor Darknet and detecting existing or new suspicious contents. These tools are designed to support the Spanish Law Enforcement Agencies (LEAs) in their surveillance of the Tor hidden services. An overview of two of our current contributions to the Tor monitoring tool is summarized in Figure \ref{fig:system_arch}.

The first contribution to the monitoring system presented in Figure \ref{fig:system_arch} was a classification module, which detects and isolates the categories of suspicious onion domains that Spanish LEAs are interested in monitoring. For this task, we used our supervised text classifier presented in \cite{al2017classifying}, which categorizes HSs into eight classes: Pornography, Cryptocurrency, Counterfeit Credit Cards, Drugs, Violence, Hacking, Counterfeit Money, and Counterfeit Personal Identification including Driving-License, Identification, and Passport.

The second module, which is also the focus of this work, addresses the problem of \textit{ranking} the hidden services that were classified as suspicious. Once they are ranked, a police officer can prioritize her/his work by focusing on the most influential HSs, i.e., those which are in the first positions of the rank.
In our previous work \cite{alnabki2019torank}, we presented a ranking algorithm, called ToRank, to sort the onion domains by analyzing the connectivity of their hyperlinks, what was a linked-based approach. In this case, sharing the same objective, we propose a more rich solution for ranking them by analyzing as well the content of the domains.

One of the difficulties we faced was how to define the influence of a given onion domain based on its ability to attract the public. In this work, we assess the attractiveness of an onion domain, and we assign accordingly to it a score that reflects its influence among the other domains with similar content. Hence, the more attractive is a domain, the higher the score it receives. Using our text classifier, the ranking module we are proposing works at category-level and detects the influential HS in each category. Therefore, this paper aims to answer the following question: \textit{What are the most attractive onion domains in a determined area of activities?}

The answer to this question could improve the capability of LEAs in keeping a close eye on the suspicious Hidden Services more influential by concentrating their efforts in monitoring them rather than the less influential ones. Moreover, in the case a Law Enforcement Agency takes a suspicious HS down, the proposed ranking module would recognize the same domain again if it still holds the same content, even if it were hosted under a different or new address. Similarly, when a new domain is introduced to the network for the first time, and it hosts suspicious content similar to an HS that was previously nominated as influential, the ranking module could capture it before becoming popular among Tor users. Therefore, LEAs will have the needed information to strike the suspicious domains preemptively earlier.

A straightforward strategy to detect the influential onion domains is to sort them by the number of clients' requests that they receive, i.e., analyzing the traffic of the network. However, the design of the Tor network is oriented to preventing this behavior \cite{torblog2009}. Chaabane et al. \cite{chaabane2010digging} conducted a deep analysis for the Tor network traffic through establishing six \textit{exit nodes} distributed over the world with the default exit policy. However, this approach can not assess the traffic of onion domains that are not reachable through these exit nodes. Furthermore, it could be risky because the Tor network users could reach any onion domain, regardless of its legality, through the IP addresses of the machines dedicated to that purpose. Biryukov et al. \cite{biryukov2014content} tried to exploit the concept of \textit{entry guard nodes} \cite{elahi2012changing} to de-anonymize clients of a Tor hidden service. However, this proposal will not be feasible as soon as the vulnerability is fixed.

Another strategy is to employ a link-based ranking algorithm such as ToRank \cite{alnabki2019torank}, PageRank \cite{PageRank1999}, Hyperlink-Induced Topic Search (HITS) \cite{HITS1999}, or Katz \cite{Katz1953}. We explored it in our previous work \cite{alnabki2019torank}, and we concluded that the main drawback of this approach lies in its dependency on hyperlinks connectivity between the onion domains \cite{bernaschi2017exploring}. Hence, if an influential but isolated domain exists in the network, this technique can not recognize it as an essential item. 

In this paper, we present an alternative approach that incorporates several features that are extracted from the HSs into a Learning to Rank (LtR) schema \cite{li2011short}. Given a list of hidden services, our model ranks onion domains based on two key steps: content feature extraction and onion domain ranking. First, we represent each onion domain by a forty elements feature vector extracted from five different resources that are: 1) the textual content of the domain, 2) the textual Named Entities (NE) in the user-visible text like products names and organizations names, 3) the HTML markup code by taking advantage of specific HTML tags, 4) the visual content like the images exposed in the domain, and finally 5) the position of the targeted onion domain with respect to the Tor network topology. Second, the extracted features are cleaned and normalized to train a ranking function using the LtR approach to rank the domains and to propose top-$k$ HSs as the most influential. 

The ranking problem addressed in this paper is close to the field of Information Retrieval (IR) but with a significant difference. Both issues will retrieve a ranked list of elements similarly to how search engines work. For example, the Google search engine considers more than $200$ factors to generate a ranked list of websites concerning a query\cite{dean_2018}. However, in the context of our problem, we do not have a search term to order the results accordingly. Instead, our objective is to rank the domains based on a virtual query: \textit{What are the most attractive onion domains in a determined area of activities?} Hence, this model adopts IR to solve the problem of ranking and detecting the most influential onion domain in the Tor network, but without having available a search term.

Nevertheless, the proposed framework is not only restricted to ranking the onion domains of the Tor network. It could be generalized and be adapted to different areas with slight modifications in the feature vector, as for example, document ranking, web pages of the Surface Web, or users in a social network, among others. Our focus on this field, based on the special sensitivity of the HS contents, motivates our application in the Tor network, and, additionally, we wanted to test the use of IR techniques for single query ranking problems.

The main contributions of this paper can be summarised in the following way.

\begin{figure*}[tp]
\centering
\includegraphics[width=\linewidth]{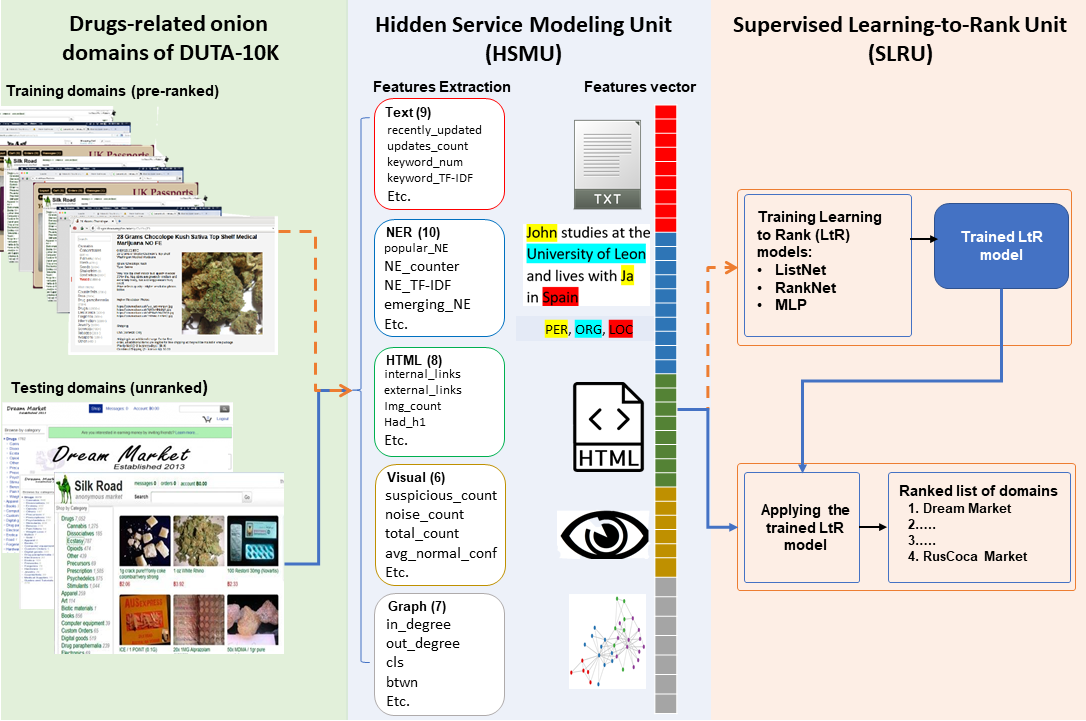}
\caption{A general view for the proposed framework for ranking and detecting the influential onion domains in the Tor network. The dashed orange arrows indicate the training pipeline of the system, while the solid blue arrows indicate the testing/production phase.}
\label{fig:framework_total}
\end{figure*}

\begin{itemize}
    \item We propose a novel framework to \textit{rank} the HSs and to \textit{detect} the most influential ones. Our strategy exploits five groups of features extracted from Tor HSs via a Hidden Service Modeling Unit (HSMU). The extracted features are used to train the Supervised Learning-to-Rank Unit (SLRU). Our approach outperforms the link-based ranking technique, such as ToRank, PageRank, HITS, and Katz, when tested on samples of onion domains related to the marketing of drugs (Fig. \ref{fig:framework_total}).

    \item We evaluated $40$ features extracted from five resources: 1) user-visible textual, 2) textual named entities, 3) the HTML markup code, 4) the visual content, and 5) features drawn from the topology of the Tor network. In particular, we address the effects of representing an onion domain by several variations of features on the ranking framework. We identify the most efficient combination of features compared to their cost of extraction in terms of the prediction time and the resources needed to build the features extraction model.
    
    \item To select the best LtR schema, we explore and compare three popular architectures: pointwise, pairwise, and listwise.
    
    \item Finally, we create a manually ranked dataset, that plays the role of ground truth for testing the models. 
\end{itemize}

The rest of the paper is organized as follows. In Section \ref{sec:state-of-the-art} we summarize the related work. After that, Section \ref{sec:approach} introduces the proposed ranking framework, including its main components. Section \ref{sec:experimental_settings} presents the experimental settings along with the configuration of the framework units. Next, Section \ref{sec:results_and_discussion} addresses a case study to test the effectiveness of the proposed framework in a real case scenario. Finally, Section \ref{sec:conclusions_and_future_work} presents the main conclusions of this work and introduces other approaches that we are planning to explore in the future.
 
\section{Related Work}
\label{sec:state-of-the-art}

There are plenty of works tackling the study of the suspicious activities that take place in the Darknet of the Tor network, as are the illicit drugs markets \cite{broseus2016studying, barratt2016everything, dolliver2015evaluating}, terrorist activities \cite{weimann2016terrorist, chen2008uncovering}, arms smuggling, and violence \cite{Fidalgo2018_PRL}, or cybercrime \cite{Ciancaglini2015, al2017classifying}

However, only a few of them have addressed the problem of analyzing the Darknet networks to detect the most influential domains. Some of them used approaches that depend on Social Networks Analysis (SNA) techniques to mine networks. Chen et al. \cite{chen2011dark} conducted a comprehensive exploration of terrorist organizations to examine the robustness of their networks against attacks. In particular, these attacks were simulated by the removal of the items featured by either their high in-degree or betweenness scores \cite{wasserman1994social}. Moreover, Al Nabki et al. \cite{alnabki2019torank} proposed an algorithm, called ToRank, to rank and detect the most influential domains in the Tor network. ToRank represents the Tor network by a directed graph of nodes and edges, and the most influential nodes are the ones which removal would reduce the connectivity among the nodes. However, the link-based approaches would fail in evaluating the isolated nodes which do not have connections to the rest of the community. Therefore, as we show in the experiments section, this approach can not surpass any of the benchmarked LtR techniques. 

Another different strategy was followed by Anwar et al. \cite{anwar2015ranking}, who presented a hybrid algorithm to detect the influential leaders of radical groups in the Darknet forums. Their proposal is based on mining the content of the user's profiles and their historical posts to extract textual features representing their radicalness. Then, they incorporate the obtained features in a customized link-based ranking algorithm, based on PageRank \cite{PageRank1999}, to build a ranked list of radically influential users.

A different perspective was followed by the study carried out by Biryukov et al. \cite{biryukov2014content}, who exploits the \textit{entry guard nodes} concept \cite{elahi2012changing} to de-anonymize clients of an onion domain in the Tor network. The popularity of onion domains in the Tor network is estimated by measuring its incoming traffic; however, this approach will not be feasible when the vulnerability is fixed.

The approach proposed in this work is entirely different and, to the best of our knowledge, is not present in the literature. We adopt a supervised framework that automatically learns how to order items following predefined ranking criteria. Concretely, we employ an LtR model to capture characteristics of a given ranked list and maps the learned rank into a new unsorted list of items.

LtR framework has been used widely in the field of IR \cite{liu2009learning, jiang2016rosf, Macdonald2013}. Li et al. \cite{li2018leveraging} proposed an algorithm to help software developers in dealing with unfamiliar Application Programming Interface (API) by offering software documentation recommendations and by training an LtR model with $22$ features extracted from four resources.
Other examples are Agichtein et al. \cite{agichtein2006improving}, who employed the RankNet algorithm to leverage search engine results by incorporating user behavior, or Wang et al. \cite{wang2017context}, who presented an LtR-based framework to rank input parameters values of online forms. They used $6$ categories of features extracted from user contexts and patterns of user inputs. Moreover, LtR has been employed for mining social networks \cite{li2016friendrank, duan2010empirical} or to detect and rank critical events in Twitter social network \cite{li2012tedas}, but none of those aforementioned works addressed the study of Tor network from a content-analysis perspective.

\section{Proposed Ranking Framework}
\label{sec:approach}
In this paper, we present a supervised framework to rank Tor Hidden Services with the purpose of measuring the influence of each domain (Fig. \ref{fig:framework_total}). Our approach has two components: 1) \textit{Hidden Service Modeling Unit (HSMU)}, which analyzes and extracts features from a given onion domain, and 2) the \textit{Supervised Learning-to-Rank Unit (SLRU)} that learns a function to order a collection of domains according to the pattern captured from previously sorted samples, a training set.

\subsection{Hidden Service Modeling Unit}
\label{lab:HSMU}
Given an onion domain $d_i \in D$, where $D$ is a set of onion domains, the HSMU analyzes $d_i$ to extract features belonging to five different categories: text, named entity, HTML, visual content, and network topology. Then, the HSMU encodes those features into numerical values that represent the HSs analyzed.

\subsubsection{Text Features} 
The set of text features involves four types of descriptors constructed from the text in $d_i$ that is visible to the user.


\vspace{.2cm}
\noindent\textbf{Date and Time:} 
A binary feature to indicate whether $d_i$ has been updated recently or not. A domain might be updated by its owner or after receiving reviews from customers concerning the offered service. We parsed the date and time patterns, and we compared them with a configurable threshold, computing this binary feature. This threshold is a particular date-time point, whereas the actions beyond it are considered obsolete.

Moreover, by counting these updates, we measured the number of recent changes that $d_i$ has received in the past. We refer to these two features by \textit{recently\_updated} and \textit{updates\_counts}, respectively.

\vspace{.2cm}
\noindent\textbf{HS Name}: 
The address of a hidden service consists of $16$ characters, randomly generated. The prefix of an HS can be customized with tools like Shallot\footnote{\url{https://github.com/katmagic/Shallot}}, allowing that the onion domain address includes words attractive to the customers. For example, a drug marketplace could add the words \textit{Cocaine} or \textit{LSD} to its HS's URL. 
Nevertheless, the customization process is extensively time-consuming; for example, customizing the first seven characters requires one day of machine-time while customizing $10$ characters takes $40$ years of processing. 

To explore this characteristic, we split the concatenated words using a probabilistic approach based on English Wikipedia unigram frequencies thanks to Wordninja tool\footnote{\url{https://github.com/keredson/wordninja}}. We obtained two features: (i) the number of human-readable words and (ii) the number of their letters. We named them as \textit{address\_words\_count} and \textit{address\_letters\_count}, respectively.

\vspace{.2cm}
\noindent\textbf{Clones Rate}: the number of onion domains that host the same content under different addresses. After reviewing Darknet Usage Addresses Text (DUTA-10K) dataset\footnote{\url{http://gvis.unileon.es/dataset/duta-darknet-usage-text-addresses/}}, we realized that some onion domains have almost the same textual content but hosted under different addresses. This feature, i.e., HS duplication, might reflect concerns of the domains' owners about their services of being taken down by authorities. The \textit{clones\_rate} feature of $d_i$ reflects the frequency of the MD5 \cite{rivest1992md5} hash code of its text.

\vspace{.2cm}
\noindent\textbf{Term Frequency-Inverse Document Frequency (TF-IDF) Vectorizer}: we used this text vectorization technique to extract and to weight domain-dependent keywords \cite{onan2016ensemble}. It allowed us to drow the following four features: 1) \textit{keyword\_num}: the number of words that are in common between the TF-IDF feature vector and the domain's words - we consider these words as the \textit{keywords} of the domain, 2) \textit{keyword\_TF-IDF\_Acc}: the accumulated TF-IDF weights of the keywords, 3) \textit{keyword\_avg\_weight}: the average weight of the keywords, and 4) \textit{keyword\_to\_total}: the number of the domain's words dived by the number of its keywords.

\subsubsection{Named Entities Features} 
A textual Named Entity (NE) refers to a real proper name of an object, including, but not limited to, persons, organizations, or locations. To extract those named entities, our previous work \cite{NABKI2019darknet} adopted a Named Entity Recognition (NER) model proposed by Aguilar et al. \cite{aguilar2017multi} to the Tor Darknet to recognize six categories of entities: persons (PER), locations (LOC), organizations (ORG), products (PRD), creative-work (CRTV), corporation (COR), and groups (GRP). We map the extracted NEs into the following five features:

\vspace{.2cm}
\noindent\textbf{NE Number}: it counts the total number of entities in $d_i$ regardless of the category; we refer to this feature by \textit{NE\_counter}.

\vspace{.2cm}
\noindent\textbf{NE Popularity}: an entity is popular if its appearance frequency value is above or equal to a specific threshold that we set to five, as is explained in Section \ref{lab:hsmu_config}. For every category identified by the NER model, we use a binary representation to capture the existence of popular entities in the domain ($1$) or ($0$) otherwise. We refer to this feature as \textit{popular\_NE\textsubscript{X}}, where $X$ is the corresponding NER category.

\vspace{.2cm}
\noindent\textbf{NE TF-IDF}: accumulates the TF-IDF weight of all the detected NE in $d_i$. This feature is denoted by \textit{NE\_TF-IDF}.

\vspace{.2cm}
\noindent\textbf{TF-IDF Popular NE}: accumulated TF-IDF weight of the popular NE, and it is named \textit{popular\_NE\_TF-IDF}.

\vspace{.2cm}
\noindent\textbf{Emerging NE}: the frequency of the emerging product entities in $d_i$. We used our previous work \cite{NABKI2017} based on K-Shell algorithm \cite{carmi2007model} and graph theory to detect these entities in the Tor Darknet. We denote this feature by \textit{emerging\_NE}.

\subsubsection{HTML Markup Features} 
Using regular expressions, we parse the HTML markup code of $d_i$ to build the following eight features:

\vspace{.2cm}
\noindent\textbf{Internal Hyperlinks}: number of unique hyperlinks that share the same domain name as $d_i$. We denote it by \textit{internal\_links}.

\vspace{.2cm}
\noindent\textbf{External Hyperlinks}: refers to the number of pages referenced by $d_i$ on Tor network or in the Surface Web. We refer to this feature by \textit{external\_links}.

\vspace{.2cm}
\noindent\textbf{Image Tag Count}: denotes the number of images referenced in $d_i$. It is calculated by counting the \textit{$<img>$} HTML tag in the HTML code of $d_i$. We denote it by \textit{img\_count}.

\vspace{.2cm}
\noindent\textbf{Login and Password}: a binary feature to indicate whether the domain needs login and password credentials or not. We use a regular expression pattern to parse such inputs. This feature is called \textit{needs\_credential}. 

\vspace{.2cm}
\noindent\textbf{Domain Title}: a binary feature that checks whether the \textit{$<title>$} HTML tag has a textual value or not. We call it \textit{has\_title}.

\vspace{.2cm}
\noindent\textbf{Domain Header}: a binary feature that analyzes if the \textit{$<H1>$} HTML tag has a header or not. We named it \textit{has\_H1}.

\vspace{.2cm}
\noindent\textbf{Title and Header TF-IDF}: an accumulation of the TF-IDF weight for the $d_i$ title and header text. It is denoted by \textit{TF-IDF\_title\_H1}.

\vspace{.2cm}
\noindent\textbf{TF-IDF Image Alternatives}: TF-IDF weight accumulation of the alternative text. Some websites use an optional property called \textit{$<alt>$} inside the image tag \textit{$<img>$} to hold a textual description for the image. This text becomes visible to the end-user to substitute the image in case it is not loaded properly. It is denoted as \textit{TF-IDF\_alt}.

\subsubsection{Visual Content Features}
The visual content could be an attractive factor for drawing the attention of end-users rather than the text, especially in the Tor HSs when the customer seeks to have a real image of the product before buying it. A suspicious services trader might incorporate real images of products to create an impression of credibility to a potential customer. However, the visual contents that are interesting for LEAs could be mixed up with other noisy contents, such as banners and images of logos. To isolate the interesting ones, we built a supervised image classifier that categorizes the visual content into nine categories, where eight of them are suspicious, and one is not. The definition of these categories is based on our previous works \cite{al2017classifying, alnabki2019torank}, where we created DUTA dataset and its extended version DUTA-10K. The image classification model was built using Transfer Learning (TL) technique \cite{lu2015transfer} on the top of a pre-trained Inception-ResNet V2 model \cite{szegedy2017inception}. The visual content feature vector has six dimensions distributed in the following manner:

\vspace{.2cm}
\noindent\textbf{Image Count}: represents the total number of images in $d_i$, both suspicious and non-suspicious images regardless of their category. Suspicious stands for images that could contain illicit content. We denote these features by \textit{total\_count}, \textit{suspicious\_count} and \textit{noise\_count}, respectively.

\vspace{.2cm}
\noindent\textbf{Average Classification Confidence}: represents the averaged confidence score of multiple images per category. These features are named \textit{avg\_suspicious\_conf} and \textit{avg\_normal\_conf}, respectively.

\vspace{.2cm}
\noindent\textbf{Majority Class}: a binary flag to indicate whether the majority of the images published in $d_i$ are suspicious or not. This flag is denoted by \textit{suspicious\_majority}.

\subsubsection{Network Structure Features}
Additionally, we modeled the Tor network by a directed graph of nodes and edges. The nodes refer to onion domains and the hyperlinks between domains are captured by the edges. This representation allowed us to built the following seven features:

\vspace{.2cm}
\noindent\textbf{In-degree}: refers to the number of onion domains pointing to the domain $d_i$. It is called \textit{in\_degree}.

\vspace{.2cm}
\noindent\textbf{Out-degree}: indicates the number of hidden services referenced by $d_i$, and it is named \textit{out\_degree}.

\vspace{.2cm}
\noindent\textbf{Centrality Measures}: for each domain $d_i$ in the Tor network graph, we evaluated three node's centrality measures: \textit{closeness}, \textit{betweenness}, and \textit{eigenvector} \cite{freeman1978centrality}. In particular, the closeness measures how short the shortest paths are from $d_i$ to all domains in the network, and it is named \textit{cls}. The betweenness measures the extent to which $d_i$ lies on paths between other domains, and it is named \textit{btwn}. Finally, the eigenvector reflects the importance of $d_i$ for the connectivity of the graph and it is denoted \textit{eigvec}.

\vspace{.2cm}
\noindent\textbf{ToRank Value}: ToRank is a link-based ranking algorithm to order the items of a given network following their centrality \cite{alnabki2019torank}. We applied ToRank to the Tor network to rank the onion domains, and we used the assigned rank as a feature of the node. Moreover, we used a binary flag to indicate whether $d_i$ is in the top-X domains of ToRank or not. We refer to those features as \textit{ToRank\_rank} and \textit{ToRank\_top\_X}, respectively.

After computing the forty features described (Table \ref{table:feature_vector}), we concatenate them to form a feature vector. However, given the variety of the scales of the features, we normalize them by removing the mean and scaling to unit variance.

\begin{table}[htp]
\label{table:feature_vector}
\caption{Summarization of the HSMU feature vector}
\resizebox{\linewidth}{!}{%
\begin{tabular}{lrll}
\hline
\multicolumn{1}{c}{\textbf{Feat. Class}} & \multicolumn{1}{c}{\textbf{\begin{tabular}[c]{@{}c@{}}Feat.\\ Count\end{tabular}}} & \multicolumn{1}{c}{\textbf{Feat. Source}} & \multicolumn{1}{c}{\textbf{Feat. Name}} \\ \hline
\multirow{4}{*}{Textual} & \multirow{4}{*}{9} & Date and Time & \begin{tabular}[c]{@{}l@{}}- recently\_updated\\ - updates\_count\end{tabular} \\
 &  & HS Name & \begin{tabular}[c]{@{}l@{}}- address\_words\_count\\ - address\_letters\_count\end{tabular} \\
 &  & Clones Rate & - clones\_rate \\
 &  & TF-IDF Vectorizer & \begin{tabular}[c]{@{}l@{}}- keyword\_num\\ - keyword\_TF-IDF\\ - keyword\_avg\_weight\\ - keyword\_to\_total\end{tabular} \\ \hline
\multirow{5}{*}{\begin{tabular}[c]{@{}l@{}}Named\\  Entities\end{tabular}} & \multirow{5}{*}{10} & NE Popularity & - popular\_NE (x) \\
 &  & Total NE Number & - NE\_counter \\
 &  & TF-IDF NE & - NE\_TF-IDF \\
 &  & TF-IDF Popular NE & - popular\_NE\_TF-IDF \\
 &  & Emerging NE & - emerging\_NE \\ \hline
\multirow{8}{*}{\begin{tabular}[c]{@{}l@{}}HTML\\  Markup\end{tabular}} & \multirow{8}{*}{8} & Internal Hyperlinks & - internal\_links \\
 &  & External Hyperlinks & - external\_links \\
 &  & Image Tag Count & -  img\_count \\
 &  & Login and Password & - needs\_credential \\
 &  & Domain Title & - has-title \\
 &  & Domain Header & - has\_H1 \\
 &  & TF-IDF Title and Header & - TF-IDF\_title\_H1 \\
 &  & \begin{tabular}[c]{@{}l@{}}TF-IDF Image\\  Alternatives\end{tabular} & - TF-IDF\_alt \\ \hline
\multirow{3}{*}{\begin{tabular}[c]{@{}l@{}}Visual \\ Content\end{tabular}} & \multirow{3}{*}{6} & Images Count & \begin{tabular}[c]{@{}l@{}}- suspicious\_count\\ - noise\_count\\ - total\_count\end{tabular} \\
 &  & \begin{tabular}[c]{@{}l@{}}Average Classification\\ Count\end{tabular} & \begin{tabular}[c]{@{}l@{}}- avg\_suspicious\_conf\\ - avg\_normal\_conf\end{tabular} \\
 &  & Majority Class & - suspicious\_majority \\ \hline
\multirow{4}{*}{\begin{tabular}[c]{@{}l@{}}Network \\ Structure\end{tabular}} & \multirow{4}{*}{7} & In-degree & - in\_degree \\
 &  & Out-degree & - out\_degree \\
 &  & Centrality  Measures & \begin{tabular}[c]{@{}l@{}}- cls (closeness)\\ - btwn (betweenness)\\ - eigvec (eigenvector)\end{tabular} \\
 &  & ToRank Value & \begin{tabular}[c]{@{}l@{}}- ToRank\_rank\\ - ToRank\_top\_X\end{tabular} \\ \hline
Total Features & 40 &  &  \\ \hline
\end{tabular}
}
\end{table}

\subsection{Supervised Learning-to-Rank Unit}
\label{lab:SLRU}
To learn the optimal order of the onion domains through their descriptors, we adopt an LtR approach adapted from the IR field.
In a traditional IR problem, a training sample is a vector of three components: the relevance judgment, which can be binary \cite{wang2017context} or with multiple levels of relevance \cite{li2011learning}, the query ID, and the feature vector that describes the ranked instance. However, our ranking system needs to answer a unique question: \textit{What are the most attractive onion domains in a determined area of activities?}, and in the context of our problem, the relevance judgment of an item is a numerical score that is calculated and assigned per item. These scores are set manually by a group of annotators that we considered as ground truth for the ranking, whereas the annotation procedure is described in Section \ref{subsec:dataset}. Hence, the input vector for the SLRU is limited to the features of $d_i$ and its relevant judgment $r_i$ score in $R$, where $R$ refers to the ground truth rank. The vector of each sample $d_i$ can be modeled as $V= <r_i, d_{i,1}, d_{i,2}, ..., d_{i,n}>$, $n \in N$, where $d_{i,n}$ is the $n_th$ feature of the domain $d_i$ and $N$ is the total number of the ranking features, i.e. $N = 40$.

Our LtR schema aims to learn a function $f$ that projects a feature vector into a rank value $(d_{i,1}, d_{i,2}, ..., d_{i,n})\xrightarrow{f} r_i$. Therefore, the goal of an LtR scheme is to obtain the optimal ranking function $f$ that ranks $D$ in a way similar to $R$. The learning loss function depends on the used LtR architecture, as explained in the following three subsections.

\subsubsection{Pointwise}
The loss function of the Pointwise approach considers only a single instance of onion domains at a time \cite{friedman2001greedy}. It is a supervised classifier/regressor that predicts a relevance score for each query domain independently. The ranking is achieved by sorting the onion domains according to yield scores. For this LtR schema, we explore the Multi-layer Perceptron (MLP) regressor \cite{ciaramita2008online}.

\subsubsection{Pairwise}
It transforms the ranking task into a pairwise classification task. In particular, the loss function takes a pair of items at a time and tries to optimize their relative positions by minimizing the number of inversions comparing to the ground truth \cite{burges2005learning}. In this work, we use the RankNet algorithm \cite{burges2005learning}, which is one of the most popular pairwise LtR schemes. 

\subsubsection{Listwise}
This approach extends the pairwise schema by looking at the entire list of samples at once \cite{xia2008listwise}. One of the most well-known listwise schemes is ListNet algorithms \cite{cao2007learning}. Given two ranked lists, the human-labeled scores and the predicted ones, the loss function minimizes the cross-entropy error between their permutation probability distributions.

\section{Experimental Settings}
\label{sec:experimental_settings}
To test our proposal, we designed an experiment to answer three research questions: 
\begin{itemize}
\item What is the most suitable LtR schema for the task of ranking the onion domains in the Tor network and for detecting the influential ones?
\item Which are the advantages of two different ranking approaches, the content-based and the link-based?
\item And what is the best combination of features for the LtR model performance?
\end{itemize}

In this section, we discuss the motivation behind these questions, describing in detail the analytical approach that we conducted, and finally, we present our findings. 

\subsection{Evaluation Measure}
The two most popular metrics for ranking Information Retrieval systems are Mean Average Precision ($MAP$) and Normalized Discounted Cumulative Gain ($NDCG$) \cite{jarvelin2000ir, christopher2008introduction}. The main difference between the two is that the $MAP$ assumes a binary relevance of an item according to a given query, while $NDCG$ allows relevance scores in the form of real numbers.
Two characteristics suggest the use of $NDCG$ to evaluate the problem addressed in this paper. Thanks to the first component of the Tor network monitoring pipeline - the classification components-, all the addressed onion domains are already relevant to the query, and second, the ground truth and the predicted rank produced by any of the previously commented LtR schemes are real numbers, not binary ones.
To obtain the $NDCG@K$ of a given query, we calculate the $DCG@K$ flowing the next formula (Eq. \ref{eq:dcg}).

\begin{equation} \label{eq:dcg}
DCG@K = G_{1} + \sum_{i=2}^{K} \frac{G_i}{log_2(i)}
\end{equation}


Where $G_1$ is the gain score at the first position in the obtained ranked list, $G_i$ is the gain score of the item $i$ in that list, and $K$ refers to the first $K$ items to calculate the $DCG$. To obtain a normalized version of $DCG@K$ is necessary to divide it by $IDCG@K$, which is the ideal $DCG@K$ sorted by the gain scores in descending order (Eq. \ref{eq:ndcg}).

\begin{equation} \label{eq:ndcg}
NDCG@K = \frac{DCG@K}{IDCG@K} 
\end{equation}

\subsection{Modules Configurations}

\subsubsection{Hardware Configurations}
Our experiments were conducted on a 2.8 GHz CPU (Intel i7) PC running Windows 10 OS with 16G of RAM. We implemented the ranking models using Python3.

\subsubsection{HSMU Configurations}
\label{lab:hsmu_config}
In the TF-IDF text vectorizer, we set the feature vector length to $10,000$ with a minimum frequency of $3$, following our previous work \cite{al2017classifying}. 
We used a NER model trained on WNUT-2017 dataset\footnote{\url{https://noisy-text.github.io/2017/emerging-rare-entities.html}}. To set the popularity threshold of the \textit{popular\_NE\textsubscript{X}} feature, we examined four values ($3, 5, 10, 15$), and we assigned it to $5$, experimentally. Additionally, we set the threshold of the \textit{recently\_updated} feature to three months earlier to the dataset scraping date.
To extract features from the HTML code, we used BeautifulSoup library\footnote{\url{https://pypi.org/project/beautifulsoup4/}}. To construct the Tor network graph, we used NetworkX\footnote{\url{https://networkx.github.io/}} library.

In the image classifier, we used the Transfer Learning (TL) technique with the pre-trained Inception-ResNet V2 model released by Google\footnote{\url{http://download.tensorflow.org/models/inception_resnet_v2_2016_08_30.tar.gz}}. The model was trained and tested with a dataset of $9,000$ and $2,700$ samples, respectively, equally distributed over nine categories. The motivation behind selecting these categories is to have the same classes as in our text classifier \cite{al2017classifying}. The dataset was collected from Google Images using a chrome plugin called \textit{Bulk Image Downloader}. Table \ref{table:image_classification_dataset} shows the image classifier performance per category.  

\begin{table}[htp]
\caption{The obtained F1 score by the image classifier over the testing set.}
\label{table:image_classification_dataset}
\centering
\resizebox{\linewidth}{!}{%
\begin{tabular}{lr}
\hline
\textbf{Category Name} & \multicolumn{1}{l}{\textbf{F1 Score (\%)}} \\ \hline
\textbf{Counterfeit Credit Cards} & 92.45 \\
\textbf{Counterfeit Money} & 96.78 \\
\textbf{Counterfeit Personal Identification} & 95.16 \\
\textbf{Cryptocurrency} & 94.60 \\
\textbf{Drugs} & 91.60 \\
\textbf{Pornography} & 98.53 \\
\textbf{Violence} & 93.80 \\
\textbf{Hacking} & 97.63 \\
\textbf{Others} & 86.78
\\ \hline
\end{tabular}
}
\end{table}

\subsubsection{SLRU Configurations}
We used the dataset described in Section \ref{subsec:dataset} to train and test the three LtR models introduced in Section \ref{lab:SLRU}. Due to the small number of samples in the drug domain, $290$, we conducted 5-fold cross-validation following recommendations from previous works \cite{cao2007learning}.  
On each iteration, three folds are used for training the ranking model, one fold for validation, and one fold for testing. For the three LtR models, the number of iterations is controlled by an early stopping framework, which is triggered when there is no remarkable change in the validation set at $NDCG@10$ \cite{lai2013sparse}.

The three LtR schemes commented in Section \ref{lab:SLRU} shares the same network structure but with different loss functions. The neural network has two layers, with $128$ and $32$ neurons, respectively. For non-linearity, a Rectifier Linear Unit (ReLU) activation function is used \cite{xu2015empirical}. To avoid overfitting, the ReLU layer is followed by a dropout layer with a value of $0.5$ \cite{hinton2002training}.

\section{Results and Discussion: Drugs Case Study}
\label{sec:results_and_discussion}

\subsection{Dataset Construction} 
\label{subsec:dataset}
Darknet Usage Text Addresses 10K (DUTA-10K) is a publicly available dataset proposed by Al-Nabki et al. \cite{alnabki2019torank} that holds more than $10K$ onion domains downloaded from the Tor network and distributed in $28$ categories. In this case study, we address the ranking of the onion domains that were classified as \textit{Drugs} in DUTA-10K. This category contains drugs-topics activities, including manufacture, cultivation, and marketing of drugs, in addition to drug forums and discussion groups. Out of $465$ drug domains in DUTA-10K, we selected only English language domains what makes a total of $290$ domains. This ranking approach could be adapted to any category of DUTA-10K, but we selected the drugs-related domains due to its popularity in the Tor network. We also want to stress that our approach is not only limited to web domains, and it could be extended to further fields such document ranking or influencers detection in social networks when a previously ranked list for training is available.


To create the content-based dataset, thirteen people, including the authors, ranked manually the $290$ drugs-related domains, to build a dataset that served as ground truth. For keeping the consistency in the ranking criteria among the annotators,  we created a unified questionnaire of $23$ subjective binary questions (Table \ref{table:questionnaire_23}) that were answered by the annotators for each domain. The ground truth is built in a pointwise manner assigning an annotator a value to each domain, coming from answering to every question with a $1$ or $0$, that corresponds to \textit{Yes} or \textit{No}, respectively. 

This process is repeated three times, assigning to each annotator a new batch of approximately $23$ domains every time, different in each iteration. Thus, each onion domain is judged three times by three different annotators, and as a result, each domain is represented by three binary vectors of answers. Following the majority voting approach, we unified these answers' vectors of every hidden service into a single vector of $23$ dimensions, that correspond with the number of questions. 
Finally, for a given domain, the vector of answers is aggregated into a single real number, a gain value, by adding up its elements, corresponding the obtained sum with the ground truth rank of that domain. The higher the gain value, the higher the rank an onion domain obtains.

\begin{table}[htp]
\centering
\caption{The binary questionnaire used to build a ground truth rank for the drugs onion domains.}

\label{table:questionnaire_23}
\resizebox{\linewidth}{!}{%
\begin{tabular}{ll}
\hline
\multicolumn{2}{c}{\textbf{Questions}} \\ \hline
Has a satisfactory FAQ? & Has a communication channel? \\
Has a professional design? & Has real images for the products? \\
Has a subjective title? & Sells between 2 to 10  products? \\
Provides safe shipping? & Domain name has a meaning? \\
Offers reward or discount? & Products majority are illegal? \\
Sell more than 10 products? & Still accessible in TOR network? \\
Shipping worldwide service? & Sells at least one popular product? \\
Reputation content? & Requires login/ registration? \\
Accepts only Cryptocurrency? & Recently updated? \\
\begin{tabular}[c]{@{}l@{}}Can customers add a review/\\feedback?\end{tabular} & \begin{tabular}[c]{@{}l@{}}Do you feel that this domain\\is trustable?\end{tabular} \\
\begin{tabular}[c]{@{}l@{}}Need text spotting for the\\ products' images\end{tabular} & \begin{tabular}[c]{@{}l@{}}Are you satisfied with the\\ products description?\end{tabular} \\
Has more than 10 sub-pages? &  \\ \hline
\end{tabular}
}
\end{table}

\subsection{Learning to Ranking Schema Selection}
\begin{figure}
\centering
  \includegraphics[width=\linewidth]{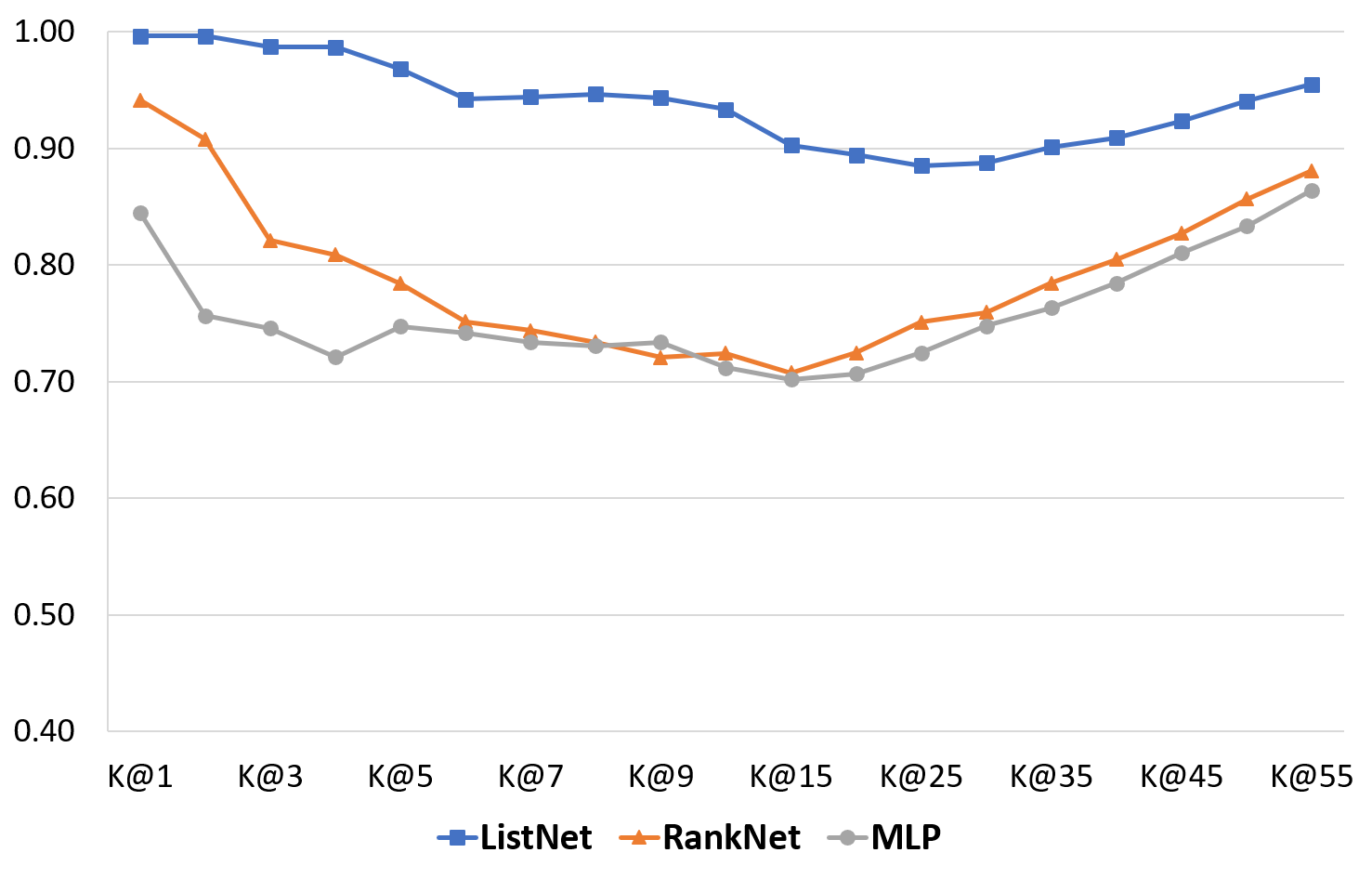}
  \caption{A comparison between three LtR algorithms against multiple values of $NDCG@K$. The horizontal axis refers to $K$ value and the vertical one indicates the $NDCG$ scores of the algorithms, obtained at each value of $K$.}
  \label{fig:ltr_3_alg_compare}
\end{figure}

In Section \ref{lab:SLRU}, we evaluated three well-known LtR schemes, namely, pointwise, pairwise, and listwise, and for each one, we explored a supervised ranking algorithm: MLP, RankNet, and ListNet, respectively. We wanted to know what is the most suitable LtR schema for the task of ranking the onion domains in the Tor network and detecting the influential ones. Fig. \ref{fig:ltr_3_alg_compare} compares the three LtR algorithms for $NDCG$ at ten different values of $K$, and given the importance of having the head of a list ranked correctly more than its tail \cite{cao2006adapting, wang2017context}, we illustrate the top-$10$ values of $K$ individually. 

Fig. \ref{fig:ltr_3_alg_compare} shows the superiority of the Listwise approach when each domain is represented by a vector of $40$ features extracted from five different kinds of resources (Section \ref{lab:HSMU}). 
The same figure shows that the $NDCG@1$ of the ListNet is equal to one, which means that during the five folds of cross-validation, the algorithm ranked correctly the first domains tested, exactly as the ground truth. At $NDCG@4$, the curve starts dropping; however, the lowest $NDCG$ value was $0.88$ for $K$ equals to $25$. As also can be seen in Fig. \ref{fig:ltr_3_alg_compare}, the pointwise approach, which is the MLP in our case,  obtained the worst performance, which agrees with the conclusion of other researchers \cite{li2011learning}.

In addition to comparing the performance in terms of the $NDCG@K$, we registered the duration of the time required to train and to test each LtR model, i.e., starting from the moment the model receives a list of domains encoded by HSMU (Section \ref{lab:HSMU}) until it produces their rank. On average, for the five-folds, the ListNet model took $8.30$ seconds for training and $0.08$ seconds for testing. The RankNet took $7.35$ seconds for training and $0.007$ seconds for testing. Finally, the fastest one was the MLP model, which took $3.34$ seconds for training and $0.0009$ seconds for testing. This comparison shows that the ListNet model is the slowest one due to the complexity of its loss function comparing to the ones of the RankNet and the MLP algorithms.

\subsection{Link-based versus Content-based Ranking}
Having two distinct ranking strategies raises a question: \textit{which is the most suitable ranking approach? Content-based or link-based?}. To answer it, we explore four link-based algorithms, namely, ToRank \cite{alnabki2019torank}, PageRank \cite{PageRank1999}, Hyperlink-Induced Topic Search (HITS) \cite{HITS1999}, and Katz \cite{Katz1953}. In particular, we compare the best LtR model of our approach, i.e., ListNet, which is considered as a supervised ranking algorithm, with the four link-based algorithms that are unsupervised. We represent the Tor network by a directed graph that consists of nodes and directed edges, where nodes represent the onion domains, and the hyperlinks between them are captured using the directed edges.

\textbf{\textit{Comparison configuration}}. Unlike our approach, the link-based algorithms do not require training data. To obtain a fair comparison, we carried out $5$-folds cross-validation with the same random seed for both ranking approaches. Then, we constructed a directed graph out of the testing nodes and applied the link-based algorithms; finally, both approaches were evaluated using the same test set. For the link-based algorithms, we evaluated several configuration parameters and selected the ones that obtained the highest $NDCG$ (Table \ref{table:algo_config}).

\begin{table}[htp]
\centering
\caption{The examined parameters for the link-based ranking algorithms. Bold numbers correspond to the selected configuration that achieved the highest $NDCG$ value.}
\label{table:algo_config}
\begin{tabular}{llr}
\hline
\hline
\textbf{Algorithm Name} & \textbf{Parameter} & \multicolumn{1}{l}{\textbf{Evaluated values}} \\ \hline
\multirow{2}{*}{\textbf{PageRank}} & alpha &0.5, 0.70, 0.75.0.80,\textbf{0.85},0.90 \\
 & max\_iter & 10, 100, \textbf{1000}, 10000 \\ \cline{2-3} 
 \multirow{2}{*}{\textbf{ToRank}} & alpha & 0.50, 0.70, 0.80, \textbf{0.90}, 1.00 \\
 & beta & 0.1, \textbf{0.2}, 0.3, 0.4, 0.5, 0.6 \\ \cline{2-3} 
\textbf{HITS} & max\_iter & 10, 100, \textbf{1000}, 10000 \\ \cline{2-3} 
\multirow{3}{*}{\textbf{Katz}} & alpha & 0.01, \textbf{0.1}, 0.2, 0.3, 0.4, 0.6, 0.9 \\
 & beta & 0.1, 0.3, 0.5, 0.7, 0.9, \textbf{1.0} \\
 & max\_iter & 10, 100, \textbf{1000}, 10000 \\ \cline{2-3} 
 \hline
\end{tabular}
\end{table}

\begin{figure}[tp]
 \includegraphics[width=\linewidth]{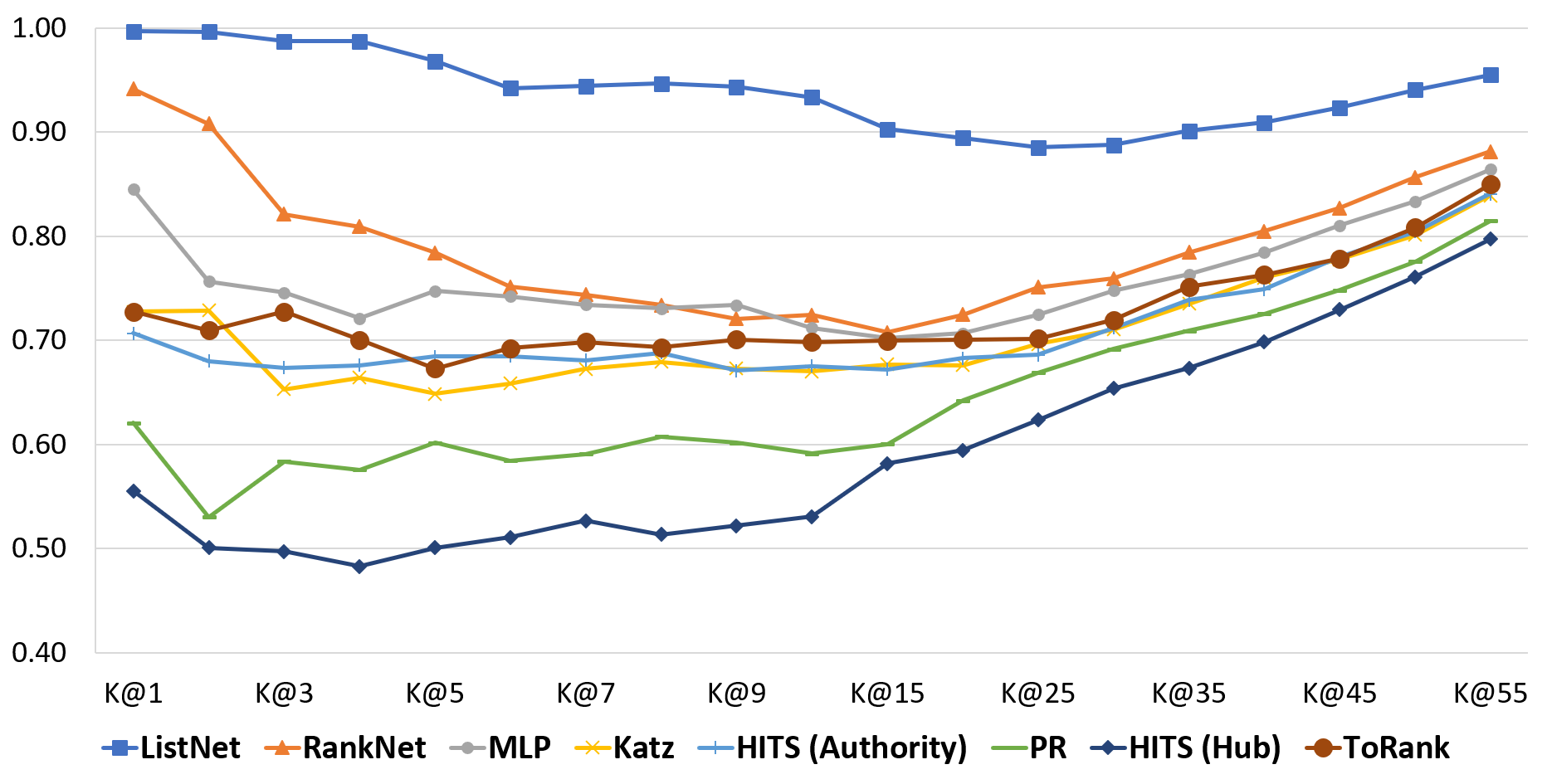}
 \caption{A comparison between the content-based versus link-based ranking algorithms with respect to multiple values of $K$. The horizontal axis refers to $K$ value and the vertical one indicates the $NDCG$ scores the algorithms obtained at each value of $K$.}
 \label{fig:content_vs_links}
\end{figure}

Fig. \ref{fig:content_vs_links} shows that ListNet highly surpasses all the link-based ranking algorithms. We observe that the weakest LtR approach, i.e., MLP, which obtained a $NDCG@10$ of $0.71$, outperforms the best link-based ranking algorithm, ToRank, which scored $NDCG@10$ of $0.69$. This result emphasizes the importance of considering the content of domains rather than their hyperlinks connectivity only.

\subsection{Feature Selection}
In the previous sections, we concluded that ListNet outperforms the RankList and MLP content-based and the four link-based algorithms when the proposed forty features represent an onion domain. However, the cost of these features varies. Some of them, such as the visual content, requires building a dedicated image classification model, while other features could be extracted merely using a regular expression. The cost is reflected in the time necessary to obtain the features and to build the ranking model, in addition to the inference time. On average, per domain, the prediction of the image classification model was the most expensive and took $109$ seconds, followed by the NER model with $22$ seconds, and then it was the text features that required $12$ seconds. Finally, the HTML and the graph features were the fastest ones to be extracted, spending $3$ and $2$ seconds, respectively.

In the following, we want to answer the question: \textit{what is the feature or combination of features that produce the best performance of the LtR model ?} To answer this question, we conducted feature analysis for the best LtR model, the ListNet one, using features and combining them from the five different resources. To discover the best combination of features for the ranking system performance, we trained and tested five ListNet models for each source of features and compared their performance.

\begin{figure*}[htb]
 \includegraphics[width=\textwidth]{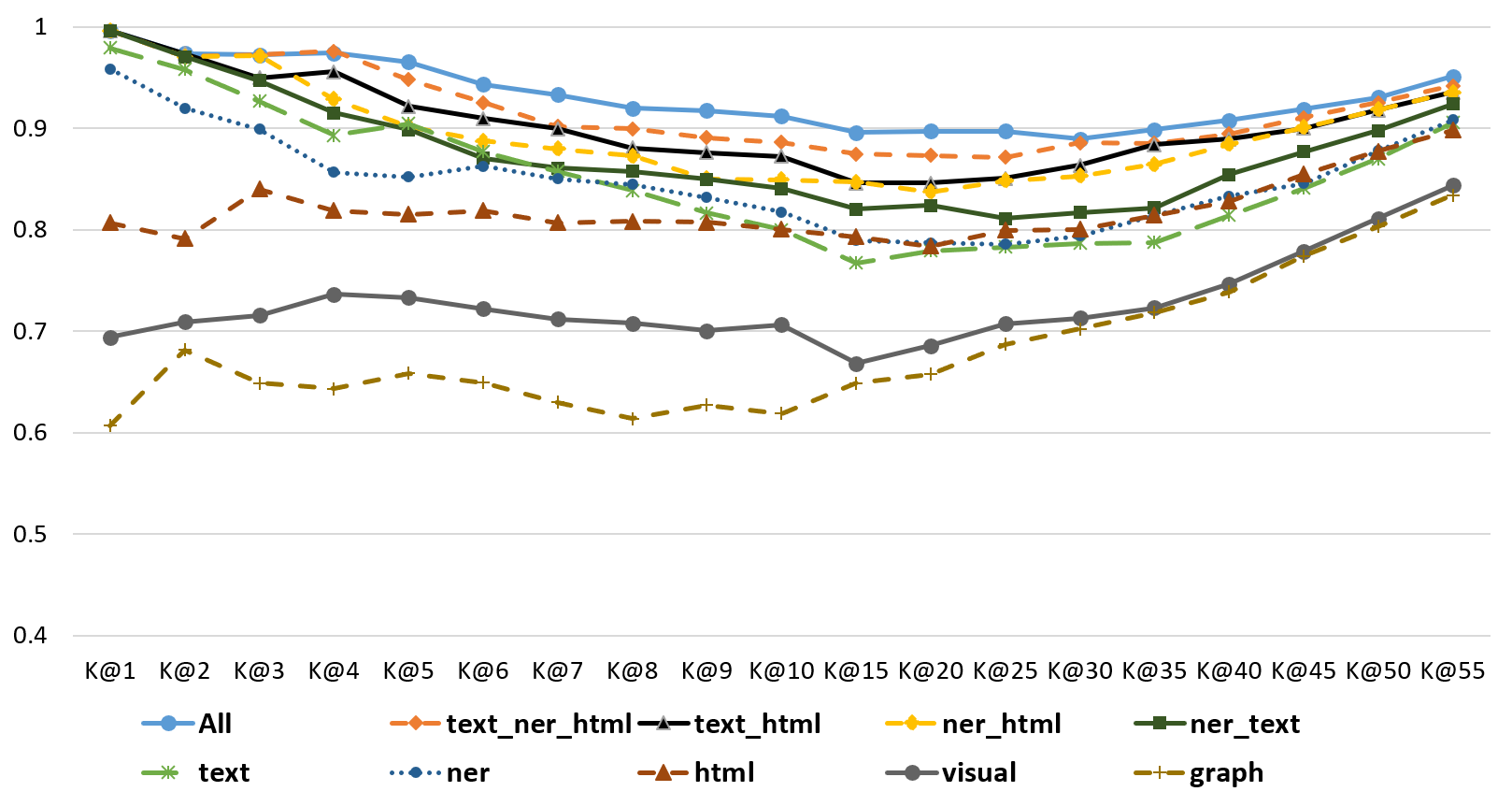}
 \caption{The effect of using different types of features along with their combinations on the ListNet ranking model. The vertical axis refers to the $NDCG$ value, while the horizontal axis denotes the value of $K$. Each curve refers to a source of features: textual (\textit{text}), featured produced by a Named Entity Recognition (\textit{NER}), HTML markup features (\textit{HTML}),  visual features (\textit{visual}), graph features (\textit{graph}), and all the features fused, denoted as (\textit{All}).}
 \label{fig:features_analysis_ndcg_listnet}
\end{figure*}

When we analyze only features coming from a single source, without combining them, it can be seen in Fig. \ref{fig:features_analysis_ndcg_listnet} that the features that are extracted only from text, denoted by \textit{text} and in clear green in this figure, achieves the highest $NDCG@5$ of $0.90$. Very close to them is the model trained using only recognized named entities (NER) features, which obtains a $NDCG@5$ of $0.85$. After that, using only features extracted from HTML, the ListNet model obtains a $NDCG@5$ of $0.81$. In contrast, the graph features obtain the lowest $NDCG@5$ of $0.65$, which indicates their weakness for ranking the onion domains, unlike the features that are extracted from the text, which have shown a significant and positive impact on the $NDCG$ metric. Hence, we conclude that the features extracted from the user-visible text are more representative comparing to the ones coming from the visual characteristics of the domain or the graph ones.

After the previous analysis, we decided to investigate the effect of combining features from different sources, to measure the impact of those combinations on the model performance. In Figure \ref{fig:features_analysis_ndcg_listnet} it can be seen that the performance increases when the top-$3$ individual features, i.e. \textit{text}, \textit{NER}, and \textit{HTML} are fused. Also, those three features combined could obtain a $NDCG$ close to the one yield by combining all the features (\textit{All}). Consequently, we could remove the graph and the visual features and keep the ranking performance relatively high and very close to the case when all the features are used.

\section{Conclusions and Future Work}
\label{sec:conclusions_and_future_work}
The Tor network is overfull with suspicious activities that LEAs are interested in monitoring. By ranking the onion domains according to their influence inside the Tor network, LEAs can prioritize the domains to leverage the monitoring process. 

In this paper, we created several ranking frameworks using Learning-to-Rank (LtR) to detect the most influential onion domains in the Tor darknet using different sources of features. Our purpose was to determine which is the best approach that allows the best performance with the lower complexity in the model creation. The proposed framework consists of two components. 1) Hidden Service Modeling Unit (HSMU), that represents an onion domain by $40$ features that are extracted from five different resources: the domain user-visible text, the HTML markup of the web page, the named entities in the domain text, the visual content, and the Tor network structure; and 2) Supervised Learning-to-Rank Unit (SLRU), that learns how to rank the domains using LtR approach. To train the LtR model, we built a manually sorted dataset of $290$ drugs-related onion domains. 

We tested the effectiveness of our framework on a manually ranked dataset of onion domains related to drug trading. We explored and evaluated three common LtR algorithms: MLP, RankNet, and ListNet, considering that the method which obtained the highest $NDCG$ is the best one. We found that the ListNet algorithm outperforms the rest of the ranking algorithms with a $NDCG@10$ of $0.95$.
Moreover, we contrasted our framework with four link-based ranking algorithms, and we observed that the MLP with a $NDCG@10$ of $0.71$, which is the worst LtR algorithm, is better than the best link-based one, ToRank, which obtained a $NDCG@10$ of $0.69$.
 
Given the superiority of the ListNet algorithm, we analyzed how the different kinds of features impact the ranking performance. We found that using only the features extracted from the user-visible textual content, including the text, the named entities, and HTML markup code, the model achieves a $NDCG@4$ of $0.97$, exactly the same as the model that uses all the features. However, at $NDCG@10$, the performance drops slightly to $0.88$ comparing to $0.91$ using the five different sources of features. Considering both the cost to obtain the features and to create the models and its score, we recommend the text-NER-HTML model because its cost is low, and its score is almost the same as the more complex approach that uses all the features.

In the future we plan to explore additional LtR methods from the listwise approach such as RankBoost \cite{freund2003efficient} and LambdaMart \cite{burges2010ranknet}. Moreover, we will explore the StarSpace algorithm \cite{wu2017starspace} that attempts to learn objects representations into a common embedding space that could be used to entities ranking and recommendation systems.
Finally, in order to ease the process of building the training dataset, both in terms of time and the number of the labeled samples, we plan to explore the Active Learning technique \cite{Jack2016}, which selects the most distinct samples to be sorted by an expert.

\ifCLASSOPTIONcompsoc
  \section*{Acknowledgments}
\else
  \section*{Acknowledgment}
\fi
This work was supported by the framework agreement between the University of Le\'on and INCIBE (Spanish National Cybersecurity Institute) under Addendum 01 and 22.

\ifCLASSOPTIONcaptionsoff
  \newpage
\fi

\bibliographystyle{IEEEtran} 
\bibliography{IEEEtran}

\end{document}